\documentclass[
]{ceurart}

\usepackage{listings}
\begin{document}

\copyrightyear{2025}
\copyrightclause{Copyright for this paper by its authors.
  Use permitted under Creative Commons License Attribution 4.0
  International (CC BY 4.0).}

\conference{BEAR 2025 - Workshop on Benefits of pErsonalization and behAvioral adaptation in assistive Robots, within IEEE RO-MAN 2025, August 25-29 2025, Eindhoven, The Netherlands}

\title{Towards Affect-Adaptive Human-Robot Interaction: \\
A Protocol for Multimodal Dataset Collection on \\
Social Anxiety}


\author[1]{Vesna Poprcova}[%
orcid=0009-0001-1504-6488,
email=v.p.poprcova@tudelft.nl,
url=https://www.tudelft.nl/en/staff/v.p.poprcova/,
]
\cormark[1]
\address[1]{Delft University of Technology,
  Delft, The Netherlands}
\address[2]{Erasmus University Rotterdam,
  Rotterdam, The Netherlands}

\author[1]{Iulia Lefter}[%
orcid=0000-0002-7243-2027,
email=i.lefter@tudelft.nl,
url=https://www.tudelft.nl/en/staff/i.lefter/
]

\author[2]{Matthias Wieser}[
orcid=0000-0002-0429-1541,
email=wieser@essb.eur.nl, 
url=https://www.eur.nl/en/people/matthias-wieser
]

\author[1]{Martijn Warnier}[
orcid=0000-0002-4682-6882,
email=m.e.warnier@tudelft.nl, 
url=https://www.tudelft.nl/staff/m.e.warnier/
]

\author[1]{Frances Brazier}[
orcid=0000-0002-7827-2351,
email=f.m.brazier@tudelft.nl, 
url=https://www.tudelft.nl/staff/f.m.brazier/
]

\cortext[1]{Corresponding author.}

\begin{abstract}
  Social anxiety is a prevalent condition that affects interpersonal interactions and social functioning. Recent advances in artificial intelligence and social robotics offer new opportunities to examine social anxiety in the human-robot interaction context. Accurate detection of affective states and behaviours associated with social anxiety requires multimodal datasets, where each signal modality provides complementary insights into its manifestations. However, such datasets remain scarce, limiting progress in both research and applications. To address this, this paper presents a protocol for multimodal dataset collection designed to reflect social anxiety in a human-robot interaction context. The dataset will consist of synchronised audio, video, and physiological recordings acquired from at least 70 participants, grouped according to their level of social anxiety, as they engage in approximately 10-minute interactive Wizard-of-Oz role-play scenarios with the Furhat social robot under controlled experimental conditions. In addition to multimodal data, the dataset will be enriched with contextual data providing deeper insight into individual variability in social anxiety responses. This work can contribute to research on affect-adaptive human-robot interaction by providing support for robust multimodal detection of social anxiety. 
\end{abstract}


\maketitle

\section{Introduction}

Over the past decade, interest in using technologies with Artificial Intelligence (AI) to improve mental healthcare has been growing. Automating tasks in an adaptive manner to save metal health professionals time and mental effort can lead to more efficient practices \cite{andriella_introduction_2025}. Recent reviews suggest that AI-driven mental health technologies currently do not fully address the challenges of detecting social anxiety, despite the promise that such technologies hold \cite{poprcova_exploring_2024, senaratne_critical_2022}. Social anxiety typically manifests in social interactions, characterised by discomfort, a desire for isolation, and a reluctance to engage with others \cite{jefferson_social_2001}. Individuals with social anxiety also frequently experience fear of both negative and positive feedback \cite{olfson_barriers_2000}. 

Interacting with robots tends to cause less tension than engaging with humans, regardless of anxiety levels \cite{nomura_people_2020}. This underscores the potential of socially assistive robots to support more comfortable interactions. Equipping these robots with the ability to accurately detect and adapt to the affective cues and behaviors associated with social anxiety is critical. Accurately detecting social anxiety from non-verbal affective cues and behaviours continues to present difficulties, although today’s AI technologies can effectively interpret audio signals \cite{senaratne_critical_2022}. AI technologies rely on data for training; however, a major obstacle is the scarcity of representative datasets that specifically focus on social anxiety \cite{poprcova_exploring_2024}.

To address these possibilities and limitations, this paper presents a protocol for collecting a multimodal human-robot interaction dataset using the Furhat robot \cite{al_moubayed_furhat_2012} in approximately 10-minute interactive Wizard-of-Oz (WoZ) \cite{dahlback_wizard_1993} role-play scenarios designed to elicit social anxiety. At least 70 participants will be selected and grouped according to their level of social anxiety, as assessed through online screening questionnaires. In addition, demographic and personality trait questionnaires will be administered to provide contextual data and enrich the dataset. Before the interaction, participants will complete additional questionnaires to assess their current state of social anxiety and affect, with a follow-up using the same questionnaires, conducted immediately after the interaction. During the interaction with the robot, multimodal data will be collected through audio and video recordings, as well as physiological signals captured with an Empatica EmbracePlus wristband \cite{noauthor_empatica_nodate}. Finally, participatory feedback will be collected at the end to explore the experience of the participants.

\section{Previous Work}

In this section, a review of existing datasets used to study social anxiety is provided, with a focus on the growing importance of multimodal data. It also emphasises the critical role of affect and contextual factors in social anxiety, and how such multimodal datasets can offer deeper insights. Additionally, it highlights the potential of robot-mediated interaction, that can provide a controlled environment to advance detection and enhance adaptive responses.

\subsection{Unimodal and Multimodal Datasets for Social Anxiety}

Only a limited number of datasets are currently partially or fully publicly available for the detection of social anxiety, and they vary in the modalities and stimulus scenarios used to induce social anxiety. Each of these datasets provides insights into the complex nature of social anxiety, with two datasets employing a multimodal approach to provide a more comprehensive understanding of social anxiety responses. These datasets are summarised in \textit{\textbf{Table 1}}.

Fathi et al. \cite{fathi_data_2020, fathi_development_2020} collected a text repository on the attributes of social anxiety, categorising data
into demographic, emotional, and physical symptoms from participants through a website. Salekin et al. \cite{salekin_weakly_2018} collected data from students of various ethnicities, selected through a screening survey. Participants were asked to deliver a 3-minute speech on their likes and dislikes about college or their hometown, followed by a self-report of their peak anxiety level experienced during the speech. In contrast, Boukhechba et al. \cite{boukhechba_demonicsalmon_2018} collected the DemonicSalmon dataset over a two-week period, combining self-reported anxiety measures with an impromptu speech stress task. This dataset employed passive sensing techniques throughout the study, collecting GPS location data, accelerometer data, and call and text logs. Senaratne et al. \cite{senaratne_multimodal_2021} collected a multimodal dataset by including young adults in two counterbalanced tasks designed to induce two common presentations of anxiety: a bug phobia task and a speech anxiety task. During the study, they used various sensors to capture participants' physiological data (e.g. electrocardiogram (ECG) and skin resistance) alongside behavioral data obtained through audio and video recordings. 

\begin{table} [h]
\caption{Summary of the Publicly Available Datasets Described in Subsection 2.1}
\centering
\begin{tabular}{ccccc} \toprule
Dataset & Dataset Size & Sample Size & Modalities & Labels \\ \midrule
Fathi et al., 2020 \cite{fathi_data_2020} & Repository of symptoms & 214 & text & SA/noSA  \\
Salekin et al., 2018 \cite{salekin_weakly_2018} & 3-minute samples per person & 101 & audio & Yes/No \\ Boukhechba et al., 2018 \cite{boukhechba_demonicsalmon_2018} & 2-weeks period & 59 & call, text, acc. & high/low \\ 
Senaratne et al., 2021 \cite{senaratne_multimodal_2021} & 3-minute samples per person & 95 & audio, video, phys. & high/low \\ \bottomrule
\end{tabular}
\end{table}

The restricted access to and scarcity of multimodal datasets on social anxiety limit the ability to perform robust analyses and develop more effective detection methods, highlighting the need for data collection that captures the full complexity of social anxiety \cite{poprcova_exploring_2024}.

\subsection{Multimodal Interaction Datasets: Affect and Contextual Factors}

Individuals with social anxiety primarily experience fear, often triggered by anticipation of or during social interactions \cite{stein_social_2008}. They also have difficulty regulating their emotions, making social situations particularly challenging to navigate \cite{stein_social_2008}. Although multimodal datasets provide valuable insights, there remains a need for targeted multimodal datasets that capture the affective nuances and contextual factors of social anxiety within interactive context. 

Multimodal affective datasets for healthcare and wellbeing are limited due to the sensitive nature of the data and the vulnerability of the populations, with most focusing on interactions with virtual agents rather than robots. SEMAINE is a large audiovisual dataset for building Sensitive Artificial Listener (SAL) agents capable of engaging users in sustained, emotionally nuanced conversations \cite{mckeown_semaine_2012}. Fournier et al. \cite{fournier_theradia_2024} collected the THERADIA WoZ corpus, a multimodal dataset for affective computing research in healthcare, based on current affective science and appraisal theories. This dataset involved a virtual assistant performing WoZ Computerized Cognitive Training (CCT). Another affective multimodal dataset was collected for the EMPATHIC project, aimed at improving seniors' wellbeing, where a virtual assistant played the role of a health professional coach while interacting with them \cite{palmero_exploring_2025}. An exploratory study in which children interacted with a Nao robot revealed that the robot-assisted approach was effective in identifying wellbeing-related concerns compared to traditional self-report methods, with gender differences observed in expressive responses related to mental wellbeing \cite{abbasi_analysing_2024}. Another exploratory analysis of a multimodal dataset conducted by Kebe et al. \cite{kebe_gestics_2024} also highlights the significance of contextual factors. Contextual factors such as gender \cite{xu_gender_2012}, cultural aspects \cite{hofmann_cultural_2010}, and personality \cite{kaplan_social_2015, norton_personality_1997} are important considerations in the study of social anxiety and need to be taken into account when collecting a dataset.

Contextually rich multimodal affective social interaction datasets can offer a better understanding of how individuals with social anxiety engage in social exchanges and provide insight into the dynamics of social anxiety during interactions. By capturing diverse signals—such as facial expressions, vocal tone, heart rate, and contextual cues—these datasets can serve as a foundation for developing accurate detection of social anxiety. This is particularly relevant in HRI settings, where social robot can use behavioral adaptation informed by such data to create a sense of affective support \cite{laban_sharing_2024}. 

\section{Methodology}

The experimental protocol for dataset collection was reviewed, and the Human Research Ethics Committee of TU Delft granted approval in June 2025. Participants will provide their informed consent for the research exploitation of their data in compliance with the European General Data Protection Regulation (GDPR) laws. This section details the methodology that will be applied to collect the dataset.

\subsection{Participants}
Participants will be recruited through a variety of channels, including emails, pamphlets with QR code, and social media groups. Participation in the study is entirely voluntary. This recruitment strategy will aim to secure a minimum of 70 participants while ensuring a more diverse and representative sample in order to minimise potential bias that may arise from relying exclusively on a homogeneous population, such as young Dutch students.

\subsection{Experimental Study Design}
\textbf{\textit{Figure 1}} illustrates the overall \textbf{methodology}, organised under two main categories: the \textbf{Experimental Study Design} and the \textbf{Experimental Protocol for Dataset Collection}. The Experimental Study Design serves as a preparatory stage and consists of two phases: detailed planning (i.e. interaction design phase) and implementation of a pilot study.

\begin{figure*} [ht]
 	\centering 
 	\includegraphics[width=0.9\textwidth]
 {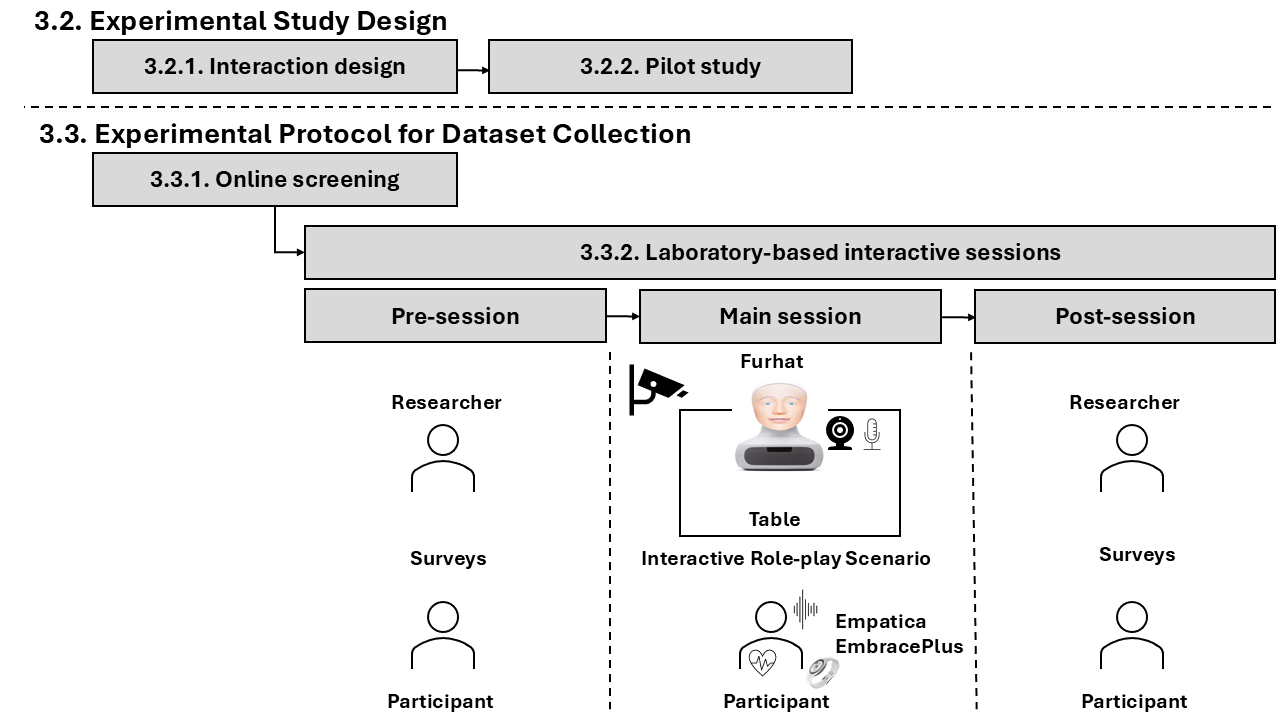}	
 	\caption{Overview of the Proposed Methodology} 
 	\label{fig_mom0}%
\end{figure*}

\subsubsection{Interaction design}
The preparation process was started with an interaction design phase, which was completed. During this phase, two psychologists were consulted to define the interaction scenarios intended to elicit social anxiety. These scenarios will later be used to implement the WoZ script for the robot dialogue. Two psychologists were invited for semi-structured, one-on-one discussions that together lasted approximately three hours. They were first introduced to the project, followed by an overview of the social robot Furhat, highlighting its characteristics and capabilities. The brainstorming part formed the core of this phase, during which both psychologists shared insights and offered constructive suggestions, with one psychologist assisting in precisely defining the scenarios.

\subsubsection{Pilot study}
To further refine the interaction design, a pilot study will involve a small number of volunteer participants to provide feedback (e.g. their thoughts and questions regarding the instructions and guidlines, any issues and suggestions related to their interaction with the robot, whether social anxiety was induced during the interaction, and the flow and realism of the conversation). The pilot study will preferably involve participants with social anxiety to ensure the collection of authentic and representative insights. Informed consent will be secured from all participants, who will be assured that their collected data will be used for research purposes and handled with strict confidentiality. The insights gathered from the pilot study will help in optimising the interaction design before the dataset collection begins. The pilot study is currently in preparation. 

\subsection{Experimental Protocol for Dataset Collection}

The Experimental Protocol for Dataset Collection will include an online screening followed by interactive sessions that will be conducted in the laboratory at TU Delft.

\subsubsection{Online screening}
The online screening will consist of online questionnaires that will include demographic information (e.g. age, sex assigned at birth, gender identity, education, and culture), and assessments of trait social anxiety, affect, and depression. In addition to culturally related demographic factors \cite{hofmann_cultural_2010}, affective processes influence social anxiety \cite{rozen_emotions_2023}, which often co-occurs with depression \cite{stein_social_2008}. Participants will be asked to complete two social anxiety questionnaires (i.e. Liebowitz Social Anxiety Scale (LSAS)\cite{fresco_liebowitz_2001} and Social Interaction Anxiety Scale (SIAS) \cite{mattick_development_1998}) to assess their level of social anxiety, together with additional questionnaires for depression (i.e. Beck Depression Inventory (BDI) \cite{beck_comparison_1996}) and trait affect (i.e. Positive and Negative Affect Schedule (PANAS) \cite{watson_development_1988}). These five questionnaires will be distributed using Qualtrics online links. 
Participants will be grouped into two groups—low and high social anxiety—based on their total scores from the social anxiety questionnaires. 

\subsubsection{Laboratory-based interactive sessions}

The interactive sessions will begin with the \textbf{pre-session} and will be carried out in the TU Delft laboratory. Upon entering the laboratory, each participant will be introduced to the robot, informed consent will be collected, and instructions will be provided on how to wear the wearable device. Each participant will then complete the state parts of the State-Trait Anxiety Inventory (STAI) \cite{zsido_development_2020} and Positive and Negative Affect Schedule (PANAS) \cite{watson_development_1988} questionnaires. 

After providing instructions, the researcher will leave the laboratory, allowing the robot to initiate the face-to-face interactive \textbf{main session} with the participant, following the setup used in \cite{ost_individual_1981} (as inlustrated in \textit{Figure 1}). Each participant will be instructed to engage in a conversation with the robot for at least ten minutes while avoiding any discussion about the recording of the conversation. The experimental study will employ a mixed factorial design, with social anxiety group as a between-subjects factor. The within-subjects factors will be robot communication style (neutral and rejecting/judgmental) and conversation depth (small talk and deep talk), both presented within a single role-play scenario. Each participant will participate in a dyadic interaction with the Furhat robot in a WoZ scripted role-play scenario designed to elicit social anxiety. The WoZ approach will be employed to control the robot’s behaviour and standardise interactions across participants, thereby ensuring consistency, reducing potential bias arising from human intervention, and helping to maintain the participant’s state as unaffected as possible by external variables.

Role-play assessments are a predominant method used in clinical settings \cite{norton_analogue_2001}. They involve the simulation of an interaction between the user and another user or a group in the clinical setting. Both standardised and individualised role-plays have been used, with advantages and disadvantages for each. Most behavioral studies of social anxiety have utilised short open-ended interactions with confederates \cite{norton_analogue_2001}. The Social Skill Behavioral Assessment System (SSBAS) \cite{norton_analogue_2001} uses an unfamiliar, opposite sex confederate and emphasises requrements of frequency with which to look at the user or engage in socially reinforcing behaviors such as nodding or smiling. The Simulated Social Interaction Test (SSIT) is another structured role-play scenario \cite{norton_analogue_2001} that assesses users' social skill and anxiety.

The design of the \textbf{role-play scenarios} in this study is further inspired and adapted based on the findings of these two studies \cite{kashdan_affective_2006, kashdan_contextual_2014}, and incorporates both \textbf{small talk and deep talk}. Kashdan et al. \cite{kashdan_contextual_2014} experimentally created social situations where there was either an opportunity for intimacy (i.e. self-disclosure deep talk) or no such opportunity (i.e. small talk). Their results showed that greater experiential avoidance during the self-disclosure conversation temporally preceded increases in social anxiety for the remainder of the interaction; no such effect was found in the small talk. Kashdan and Roberts \cite{kashdan_affective_2006} also examined the roles of trait curiosity and social anxiety. Similar work was designed to examine social anxiety in a meaningful context by evoking small talk and the reciprocal exchange of gradually escalating self-disclosure deep talk \cite{de_mooij_what_2023}. High socially anxious individuals appeared to be particularly sensitive to the aversive qualities of small talk or boring interactions. Taken together, these findings help to explain how people vary in terms of their affective reactions to interpersonal interactions, predicting differential subjective experiences within the same social interaction, and supporting the dual-process model of optimal stimulation. The role-play scenarios will incorporate both small talk and deep talk, and will be designed to include a subset of questions adapted from Aron et al. \cite{aron_experimental_1997} and further refined based on topics commonly addressed in social anxiety assessment questionnaires.

In addition to the nature of the role-play, the behavioral observation—as described by Glass and Ankorf \cite{glass_behavioral_1989}—is also characterised by aspects such as the type of interaction, the interaction partner (i.e. the robot), and the specific level of behaviour. The robot will adapt its behaviour based on a specific \textbf{communication style} designed to elicit social anxiety, consisting of \textbf{neutrality and social rejection/judgment}. Social rejection is a kind of negative social evaluation that elicits negative feelings and is considered undesirable \cite{blackhart_rejection_2009}. Rejection sensitivity is associated with social anxiety \cite{berenson_rejection_2009,gao_associations_2017}. According to the responsive theory of social exclusion, there are three options to experimentally induce social rejection: explicit rejection (i.e. clearly stating no), ostracism (i.e. ignoring), and ambiguous rejection (i.e. being unclear) actively or passively delivered \cite{freedman_softening_2016}. Social rejection can be experimentally induced to make participants feel social anxiety during interaction, using the following stimulus words: despised, hostile, immature, silly, foolish, anguish, offended (e.g. Emotional Stroop task \cite{grant_attentional_2006}). Robot anthropomorphism and heightened social presence may be perceived as more judgmental \cite{holthower_robots_2023}. Being judgmental involves the disposition to give attention to the negative features of others \cite{dake_being_2024}. It can be seen as an overlap of interpretive ungenerosity and non-acceptance \cite{watson_standing_2012}. Research exploring similar negative contexts includes robot impoliteness through phrases \cite{rea_is_2021} and robot verbal or physical punishment \cite{jois_what_2021}. The expressive capabilities of the Furhat robot will allow to operationalise these parameters and support the controlled elicitation of social anxiety by adapting its behavior—starting with a neutral expression, shifting to a more rejecting or judgmental responses during the conversation, and returning to neutrality by the end of the interaction—while leading the interaction. Appendix \ref{appendix:a} presents an example of an interaction scenario.

Following the main session, in the \textbf{post-session}, participants’ state social anxiety and momentary affect will be assessed using the same questionnaires previously administered. Personality traits, which also play a role in social anxiety \cite{kaplan_social_2015}, will be measured using the BFI-10 \cite{rammstedt_measuring_2007} (i.e. extraversion, agreeableness, conscientiousness, emotional stability, and openness). 
Subsequently,  acceptance and attitudes will be explored to analyse participants’ impressions and overall experience. At the conclusion of the interactive sessions, participants will be debriefed on the design and protocol of the experimental study for dataset collection.

\subsection{Data Collection}
This multimodal dataset will include synchronised audio, video, and physiological data collected from various sources. Experiments will be conducted to support modelling using video, audio, and physiological modalities, with a particular focus on multimodal fusion for the detection of social anxiety. Additionally, collecting contextual metadata will enable to study how different individual factors may influence social anxiety and the dynamics of the human-robot interaction.

Self-reported information will be used to collect contextual metadata, and an anonymisation process will be implemented to protect participants' identities while preserving the richness of the data. Audio and video recordings will be captured from the robot's perspective, while physiological data will be collected from the participant.

\section{Conclusion}

This research contributes to the growing body of work in affective computing for human-robot interaction in healthcare settings by providing a publicly available multimodal dataset. This dataset can offer a foundation to support the combined development of realistic multimodal behavioural modelling and generation, with particular relevance for adaptive, robot-assisted sensing and interventions targeting individuals with social anxiety. By enabling the modelling of affective states and behaviours using a combination of different modalities and multimodal fusion techniques, this dataset contributes to enhancing the detection of social anxiety. Building on this foundation, future research could explore the potential of adaptive robots to mediate social interactions in group settings (i.e. often a challenge for individuals with social anxiety), thereby providing additional support for existing findings that such robots can alleviate anxiety symptoms \cite{nomura_people_2020} and serve as effective therapeutic tools.

The study design protocol presented in this paper details the methodology for collecting this dataset, which consists of multimodal recordings of participants interacting with a WoZ-controlled Furhat robot. The interactions are carefully designed to elicit affective states and behaviours associated with social anxiety, reflecting the multidisciplinary nature of this research.

Contextual data can allow examination of social anxiety in relation to individual dimensions, such as personality and gender. Additionally, the use of two distinct robot communication styles—a rejecting/judgmental robot that expresses negative affective cues and a neutral robot endowed with warm affective expressiveness—can facilitate the investigation of how these interactional factors influence social anxiety. By considering these factors, a deeper understanding of the complexities that shape the affect-adaptive human-robot interaction can be attained.

\section*{Declaration on Generative AI}
  The author(s) have not employed any Generative AI tools.

\printbibliography
\newpage
\appendix

\section{Interaction Scenario Example}
\label{appendix:a}

Background: Ana is a socially anxious student who becomes very nervous when having casual conversations with others, including people she knows fairly well. She feels much less anxious if the conversation is about something specific, such as talking about a project. Ana does not know how to make small talk. One such situation is Friday afternoon get-together at a local bar with other students. She wants to go regularly, but had only gone twice in the last year. Once it went well because she ended up talking with someone about her project the whole time. The second time, however, everyone was talking about more personal topics. That time Ana became anxious and left early. \\

Ana's possible anticipated answers: 
\begin{itemize}
    \item I don't know how to make small talk
    \item I will become very nervous
    \item I won't have anything to say \\
\end{itemize}

The Furhat robot will use several different disputing questions to challenge these thoughts (i.e. elicit social anxiety), which is addressed in the example exposure role-play scenario presented in \textit{Table 2}.

\begin{table} [h]
\caption{HRI Exposure Role-play Scenario \cite{hope_managing_2019}}
\centering
\begin{tabular}{ccc} \toprule
Participant/Robot & Gesture & Content \\ 
\hline
 Starting Conversation & & \\
\hline
Robot & Smile & Hello. \\
Participant & & Hello. \\
Robot & Smile & I am Furhat. What is your name? \\ 
Participant & & My name is Ana. \\
Robot & Smile & Nice to meet you Ana. \\
\hline
 Small Talk & & \\
 \hline
Robot & Neutral Expression & How are you today? \\
Participant & & I am fine, thanks. How are you? \\
Robot & Frown & I would like to focus on you today. Do you have a problem \\
  & & talking with others? \\
Participant & & Yes, kind of. \\
\hline
 Deep Talk & & \\
 \hline
 Robot & Frown & Really? Can you remember a time when you said something \\
  & & embarrassing in a conversation? \\
Participant & & ... \\
Robot & Smile & What is the likelihod of this happening? \\
 & & If you imagine it, how would you prepare? \\
Participant & & ... \\
Robot & Frown & Doesn’t that sound contradictory? Please explain. \\
Participant & & ... \\
\hline
 Ending Conversation & & \\
 \hline
Robot & Smile & I understand now, apologies. \\ 
 & & It was nice to meet you, Ana. I hope to talk with you again soon. \\
\bottomrule
\end{tabular}
\end{table}

\end{document}